\documentclass[conference]{IEEEtran}
\IEEEoverridecommandlockouts
\usepackage{amsmath,amssymb,amsfonts}
\usepackage{algorithmic}
\usepackage{graphicx}
\usepackage{textcomp}
\usepackage{xcolor}
\usepackage[utf8]{inputenc} % allow utf-8 input
\usepackage[T1]{fontenc}    % use 8-bit T1 fonts
\usepackage{hyperref}       % hyperlinks
\usepackage{url}            % simple URL typesetting
\usepackage{booktabs}       % professional-quality tables
\usepackage{nicefrac}       % compact symbols for 1/2, etc.
\usepackage{microtype}      % microtypography
\usepackage{adjustbox}
\usepackage{multirow}
\usepackage{hhline}
\usepackage{biblatex}
\usepackage{wrapfig}
\usepackage{caption}
\usepackage{subcaption}
\usepackage{amsmath}
\usepackage{hyperref}

\def\BibTeX{{\rm B\kern-.05em{\sc i\kern-.025em b}\kern-.08em
    T\kern-.1667em\lower.7ex\hbox{E}\kern-.125emX}}
\begin{document}

\title{IOR: Inversed Objects Replay for Incremental Object Detection
\thanks{This work is partially supported by the Beijing Natural Science Foundation under grant (4244098), the National Natural Science Foundation of China (62476264).}
\thanks{Accepted in IEEE International Conference on Acoustics, Speech and Signal Processing, 2025 (ICASSP'25) }
}

% \and
% \and
% \and
% \and
% \and

\author{
	\IEEEauthorblockN{
		Zijia An\IEEEauthorrefmark{2}\IEEEauthorrefmark{3}, 
		Boyu Diao\IEEEauthorrefmark{2}\IEEEauthorrefmark{3}\IEEEauthorrefmark{1}
        \thanks{\IEEEauthorrefmark{1} is corresponding author.}, 
		Libo Huang\IEEEauthorrefmark{2}, 
		Ruiqi Liu\IEEEauthorrefmark{2}\IEEEauthorrefmark{3},
            Zhulin An\IEEEauthorrefmark{2}\IEEEauthorrefmark{3},
            Yongjun Xu\IEEEauthorrefmark{2}\IEEEauthorrefmark{3}} 
	
	\IEEEauthorblockA{\IEEEauthorrefmark{2}Institute of Computing Technology, Chinese Academy of Sciences\\
    \IEEEauthorrefmark{3}University of Chinese Academy of Sciences, Beijing, China\\
    Email: \{anzijia23p, diaoboyu2012, anzhulin, xyj\}@ict.ac.cn, www.huanglibo@gmail.com, liuruiqi23@mails.ucas.ac.cn}}

\maketitle

\begin{abstract}

Existing Incremental Object Detection (IOD) methods partially alleviate catastrophic forgetting when incrementally detecting new objects in real-world scenarios. However, many of these methods rely on the assumption that unlabeled old-class objects may co-occur with labeled new-class objects in the incremental data. When unlabeled old-class objects are absent, the performance of existing methods tends to degrade. The absence can be mitigated by generating old-class samples, but it incurs high costs. This paper argues that previous generation-based IOD suffers from redundancy, both in the use of generative models, which require additional training and storage, and in the overproduction of generated samples, many of which do not contribute significantly to performance improvements. To eliminate the redundancy, we propose Inversed Objects Replay (IOR). Specifically, we generate old-class samples by inversing the original detectors, thus eliminating the necessity of training and storing additional generative models. We propose augmented replay to reuse the objects in generated samples, reducing redundant generations. Moreover, we propose high-value knowledge distillation focusing on the positions of old-class objects overwhelmed by the background, which transfers the knowledge to the incremental detector. Extensive experiments conducted on MS COCO 2017 demonstrate that our method can efficiently improve detection performance in IOD scenarios with the absence of old-class objects. The code is available at \url{https://github.com/JiaJia075/IOR}.
\end{abstract}

\section{INTRODUCTION}
\label{sec:intro}

Incremental learning aims to address the challenge of catastrophic forgetting \cite{van2019three} that occurs when learning from dynamic data distributions. Typically, the process of learning new distributions leads to a reduction in the model's ability to retain knowledge of previously learned distributions \cite{wang2024comprehensive}. In incremental object detection (IOD), a specific scenario exists where images contain both unlabeled old-class objects and labeled new-class objects (co-occurrence). In this scenario, the label's distribution is dynamic, while the object's distribution remains static. Some existing methods address this scenario by leveraging unlabeled old-class objects in the images. However, the absence of old-class objects in images (non co-occurrence) is more general in real-world applications. In this scenario, the performance of IOD suffers performance degradation due to the lack of old-class information. Therefore, it is crucial to study IOD under non co-occurrence scenarios.

Existing Incremental learning methods can be classified into distillation-based \cite{shmelkov2017incremental,li2019rilod,peng2021sid,feng2022overcoming}, parameter-restriction-based \cite{liu2023augmented}, and replay-based \cite{kim2024class, liucontinual}. Most IOD methods are distillation-based, transferring the old-class knowledge from the original to the incremental detector using distillation. However, the performance of such methods degrades significantly in non co-occurrence scenarios \cite{yang2023pseudo}. Previous works \cite{yang2023pseudo,kim2024sddgr} introduce generative models to generate samples of absent classes, thereby ensuring knowledge transfer. However, the additional generators and generation processes introduced by these methods result in high computational costs. Although some methods are developed to effectively reduce computational complexity \cite{dai2024sketch, liu2024resource, yang2024clip, chen2024scp, chang2023hdsuper}, these methods still face challenges when deployed in resource-constrained real-world scenarios.

\begin{figure}[tb]
    % \vspace{-5mm}
  \centering
  \includegraphics[width=\columnwidth]{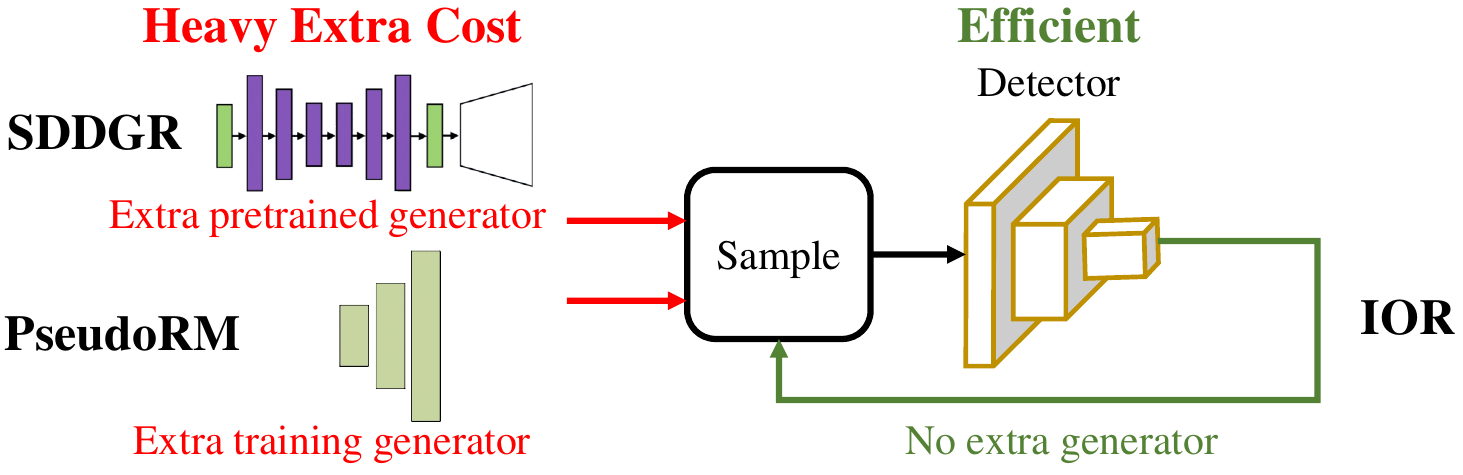}
  \caption{Compared with other methods (SDDGR\cite{kim2024sddgr} and PseudoRM\cite{yang2023pseudo}), IOR generates old-class objects without extra generators, efficiently addressing the absence of old-class objects in non co-occurrence scenarios.}
      \vspace{-7mm}
  \label{fig: IOR vs. Other Method}
\end{figure}

 \begin{wrapfigure}{r}{0.25\textwidth} % {l} 左侧，{r} 右侧
  \centering
\vspace{-3mm}
  \includegraphics[width=0.5\columnwidth]{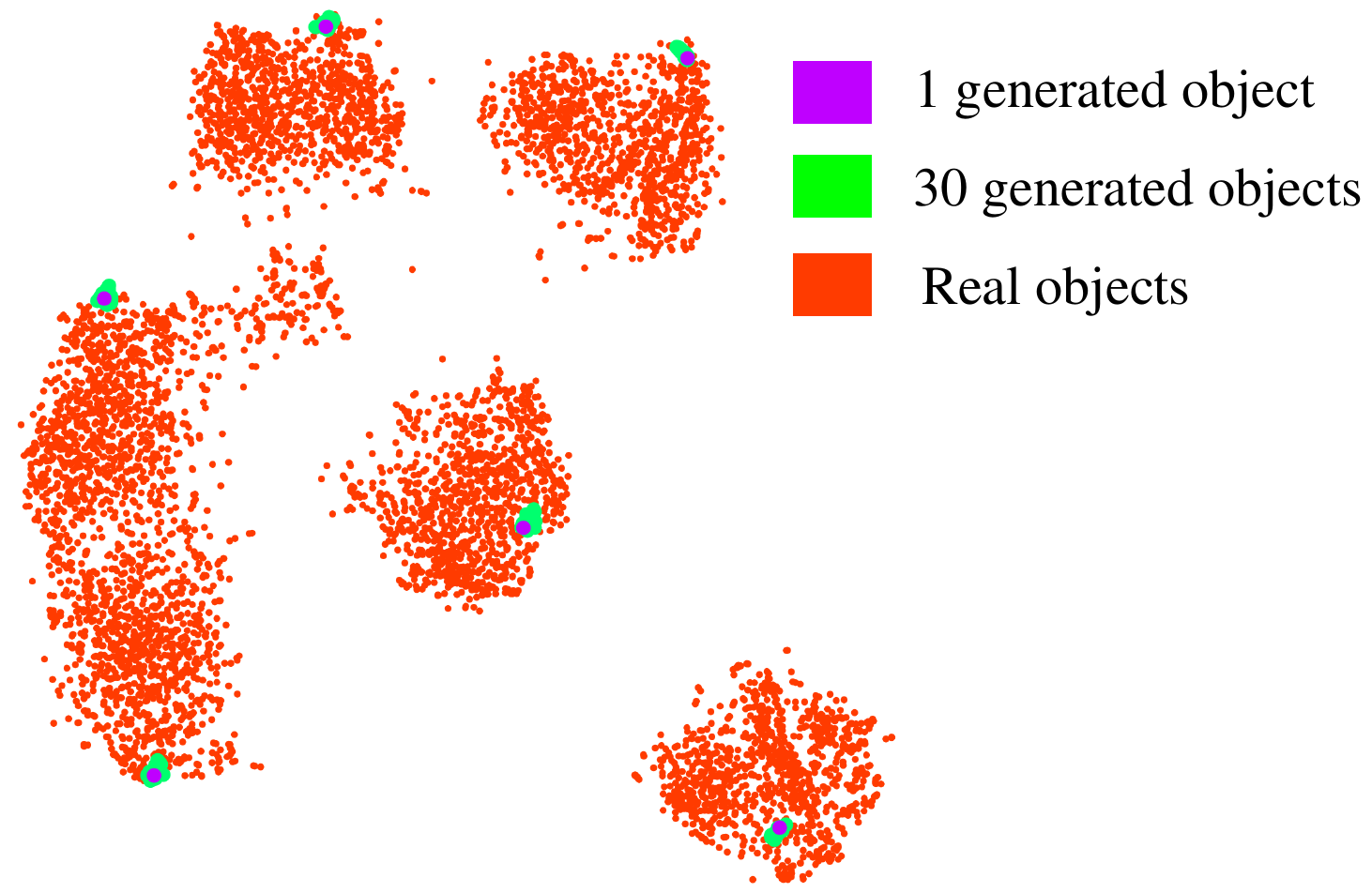}
  \vspace{-6mm}
  \caption{Visualization results for generating 1 vs. 30 objects.}
  \vspace{-3mm}
  \label{fig: 1vs30}
\end{wrapfigure}

Generative models attempt to align generated data distribution as closely as possible with real-data distribution. However, detectors only focus on certain distinctive features in the real data. The additional generative models may fail to capture the distinctive features of the detector concerns \cite{huang2024etag}. Compared with these methods, inversing \cite{yin2020dreaming} the detector to generate samples not only captures distinctive features but also eliminates the requirement for additional generative models, as shown in Fig.\ref{fig: IOR vs. Other Method}. However, we observe redundancy in inversing, as the features of the inversed samples tend to be consistent. To illustrate this phenomenon, we use t-sne \cite{van2008visualizing} to visualize the positions of the generated objects' embedding features. As shown in Fig.\ref{fig: 1vs30}, the locations of the generated objects are highly concentrated. We believe that generating only a small number of representative samples is sufficient to reflect the features of the detector concern. 
% To sum up, We eliminate the redundancy of generative models by inversing the detector and generating a small number of representative samples.

\begin{figure*}[htb]
    % \vspace{-7mm}
  \centering
  \includegraphics[width=1.4\columnwidth]{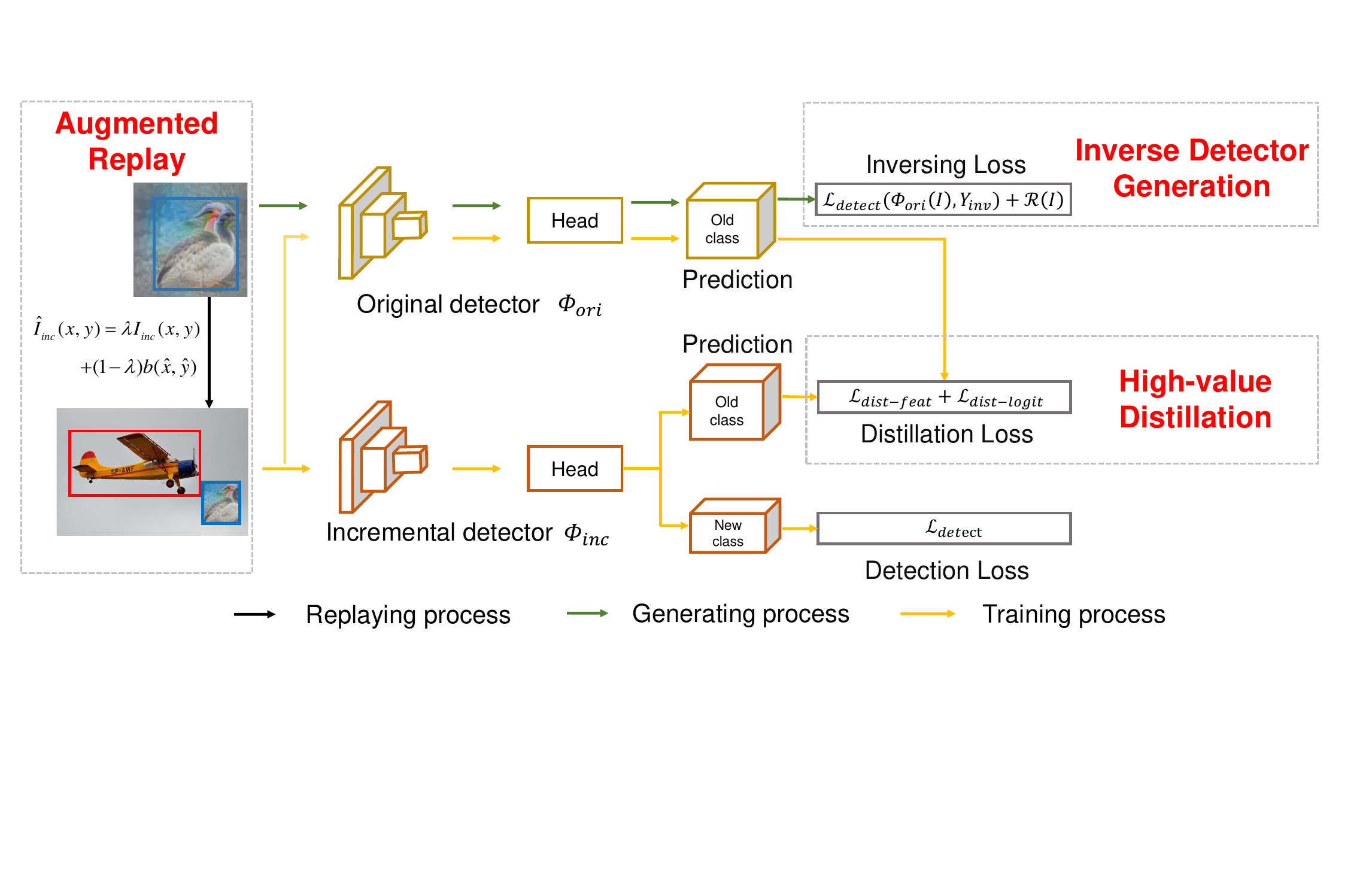}
      \vspace{-2mm}
  \caption{Framework of Inversed Objects Replay for incremental object detection.}
      \vspace{-6mm}
  \label{fig: Overall structure}
\end{figure*}

To tackle the above problems, we propose the Inversed Objects Replay (IOR). We argue that the extra cost stems from the redundancy of generative models and sample generations. We exclude redundant generative models by inversing the original detector, eliminating the necessity of training or saving generative models. We alleviate redundant sample generations by employing augmented replay, which facilitates the reuse of objects from the generated samples and thus reduces the need to generate massive samples. To effectively utilize the generated objects, we distill incremental data with replayed objects. However, the generated objects are overwhelmed by the background, leading to ineffective distillation. Therefore, we propose high-value knowledge distillation, focusing on distilling outputs relevant to old-class objects.

Our contributions can be summarized as follows: 1) To compensate for the absent objects at low cost, we inverse the original detector to generate old-class samples and reuse the objects in generated samples by augmented replay. To our knowledge, this is the first work to complement absent objects by inversing the detector in the IOD. 2) To effectively utilize the generated objects, we propose high-value distillation to focus on the objects overwhelmed by the background. 

\section{METHODS}

\subsection{Overview}

The purpose of IOR is to generate old-class objects at a lower cost and compensate for the performance degradation caused by the absence of old-class objects in non co-occurrence scenarios. Fig. \ref{fig: Overall structure} illustrates the entire framework. We generate old-class objects by inversing the original detectors to eliminate the necessity of training and storing additional generative models. Moreover, We augmently replay the generated old-class objects, thus reducing the requirement for generating objects. Since the background overwhelms the replayed objects, we propose high-value distillation to mitigate the background interference.

\subsection{Inverse Detector Generation}

We generate old-class objects by inversing the detector to eliminate the extra generative model. Given an input generated images ${I_{inv}}\in\mathbb{R}{^{C \times W \times H}}$ and the original detector ${\Phi _{ori}}$, we formulate the process of inversing the original detector as a minimization problem that each pixel is initialized from a random noise ${I_{c,w,h}} \sim N(0,1)$ and optimizes:
\begin{equation}
\label{eq1}
{I_{inv}} = \mathop {\min }\limits_I {\mathcal{L}_{detect}}({\Phi _{ori}}(I),{Y_{inv}})
\end{equation}
where ${\mathcal{L}_{detect}}$ is the loss function between the prediction of the original detector and the sampled label $Y_{inv}$. ${\mathcal{L}_{detect}}$ is the same as the detection loss of the selected detector and is responsible for determining the category and position of the generated object in ${I_{inv}}$. The sampled label ${Y_{inv}}$ consists of five parameters, an object class $cls$ and four object bounding coordinates $x$, $y$, $w$, and $h$.

To sample a more realistic label ${Y_{inv}}$, we additionally preserve the histograms of the real object's aspect ratios of each class. When inversing the detector, each image ${I_{inv}}$ is assigned a sampled label ${Y_{inv}}$. In label ${Y_{inv}}$, $cls$ is the desired class; the position $x$, $y$, $w$ is determined by a uniform distribution; the position $h = w \times ratio$, where $ratio$ is sampling from the preserved histogram. In this way, we can ensure that the generated object has a consistent aspect ratio with the real object, thus generating the object more accurately.

To make the generated images consistent with the real-data distribution, we use $\mathcal{R}(I)$ to regularize the optimization process. $\mathcal{R}(I)$ consists of two parts, a prior term ${\mathcal{R}_{prior}}$ that restricts image priors, and a regularization term ${\mathcal{R}_{BN}}$ that regularizes feature map's statistics \cite{yin2020dreaming}:
\begin{equation}
\mathcal{R}(I) = {\mathcal{R}_{prior}}(I) + {\mathcal{R}_{BN}}(I)
\label{eq2}
\end{equation}
The combination of ${\mathcal{R}_{prior}}$ and ${\mathcal{R}_{BN}}$ pushes the distribution of generated images closer to real images. The total inversing process can be expressed as:
\begin{equation}
\label{eq3}
{I_{inv}} = \mathop {\min }\limits_I {\mathcal{L}_{detect}}({\Phi _{ori}}(I),{Y_{inv}})+\mathcal{R}(I)
\end{equation}

\begin{figure}[htb]
    % \vspace{-5mm}
  \centering
  \includegraphics[width=0.65\columnwidth]{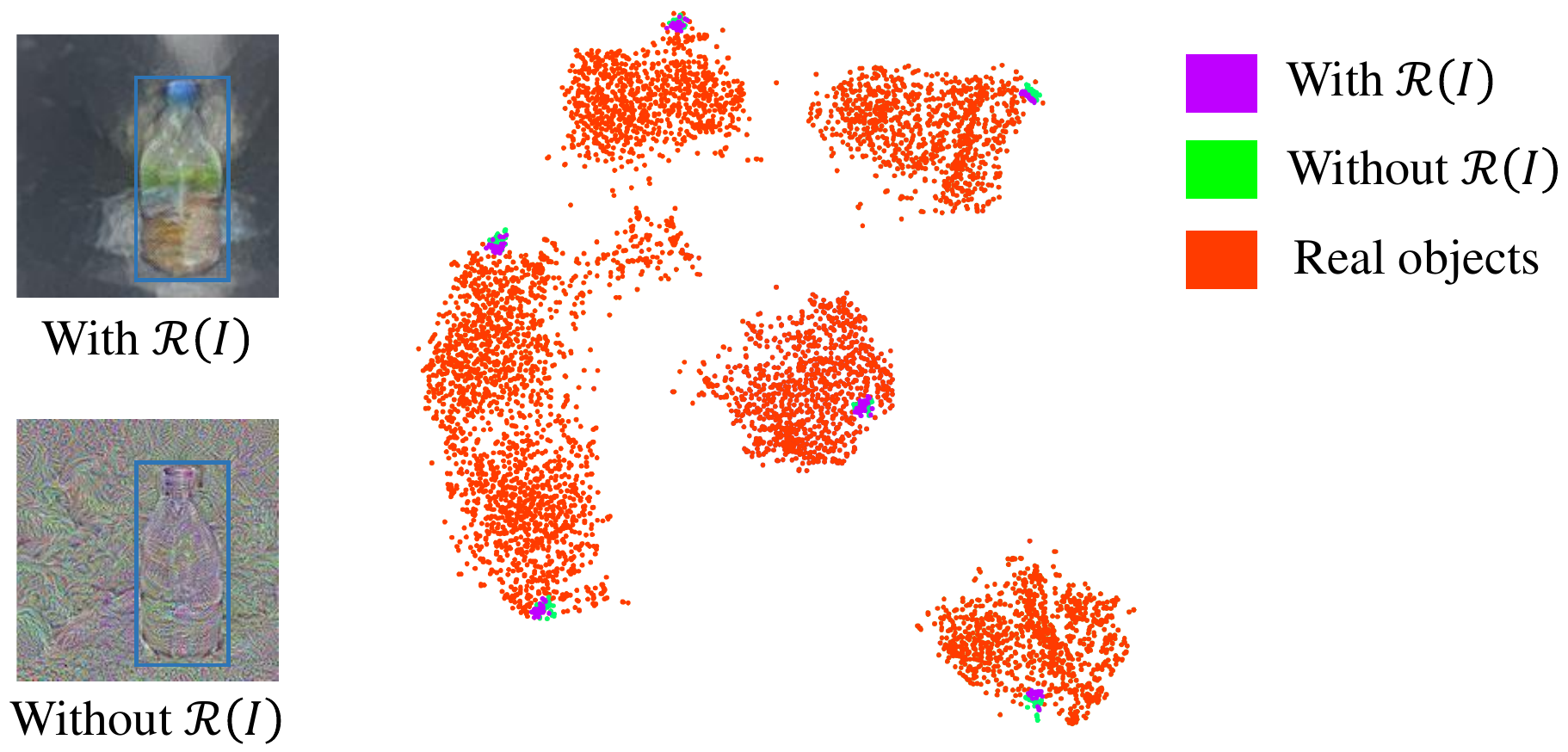}
  \caption{Comparison of visualization results with and without $\mathcal{R}(I)$ for generated object.}
      \vspace{-5mm}
  \label{fig: R(I)}
\end{figure}

It is worth noting that the increased realism of the generated objects hardly leads to increased IOD accuracy (see Section \ref{sec:Analysis and Ablation Study}). To explain this phenomenon, we use t-sne \cite{van2008visualizing} to visualize the changes in the feature distribution with and without $\mathcal{R}(I)$. In Fig.\ref{fig: R(I)}, the position of the generated objects in the embedding space remains almost unchanged, whether $\mathcal{R}(I)$ is applied or not. Therefore, $\mathcal{R}(I)$ cannot enrich the diversity of the object's features.

\subsection{Augmented Replay}

We reduce the redundancy in sample generations by reusing the generated objects. In this way, generating only a few representative objects can achieve satisfactory accuracy.

We crop the objects' bounding box $b$ from the generated images ${I_{inv}}$ by the sampled label ${Y_{inv}}$ and repeatedly place them in the incremental data. Given an image $I_{inc}$ from incremental datasets with several ground-truth bounding boxes $g$ and a cropped bounding box $b$. We assign a random location in image $I_{inc}$ for cropped bounding box $b$. Motivated by \cite{zhang2017mixup}, we mix $b$ with $I_{inc}$ to create a new image $\hat I_{inc}$. For each pixel $(x,y)$ in $\hat I_{inc}$, the mixed pixel value is computed by:
\begin{equation}
\hat I_{inc}(x, y) = 
\begin{cases} 
\lambda I_{inc}(x, y) + (1 - \lambda) b(\hat{x}, \hat{y}), & \text{if } g \cup b \leq thr \\
I_{inc}(x, y), & \text{otherwise}
\end{cases}
\label{eq4}
\end{equation}
where mixing coefficient $\lambda$ is in the range $[0,1]$ and is sampled from the Beta distribution; $b(\hat x,\hat y)$ is a cropped bounding box with a randomly assigned location; $g \cup b$ is the intersection over union (IOU) between each $g$ and $b$; $thr$ is a threshold value. If the maximum IOU between each $g$ and $b$ is less than or equal to $thr$, then the original pixel value $\hat I_{inc}(x,y)$ is a mixture of the pixel value $I_{inc}(x,y)$ and the corresponding pixel value in the cropped bounding box $b$. If the maximum IOU between a given $g$ and $b$ is greater than $thr$, then the location of the $b(\hat x,\hat y)$ requires reassignment. To reuse the generated objects better, multiple instances of $b$ can be replayed within an image $I_{inc}$, forming the set of replayed bounding boxes $B_{r}$.

\subsection{High-value Distillation}
In the feature and logit layers of the detector GFLV1  \cite{li2020generalized}, extensive background overwhelms the objects. Therefore, distilling the entire feature and logit layers suffers from background interference. To address this, we expect to distill only high-value positions in the feature and logit layers, i.e., generated objects and unlabeled real objects. The overall learning target of the incremental detector is defined as:
\begin{equation}
{\mathcal{L}_{total}} = {\mathcal{L}_{detect}} + {\mathcal{L}_{dist-feat}} + {\mathcal{L}_{dist-logit}}
\label{eq5}
\end{equation}
where the loss term ${\mathcal{L}_{detect}}$ is the detector's classification and regression loss to train incremental detectors for detecting new-class objects. The second loss term ${\mathcal{L}_{dist-feat}}$ is the distillation loss for the last feature layers at the classification head. The third loss term ${\mathcal{L}_{dist-logit}}$ is the distillation loss for the logit layers at the classification and regression head. 

\textbf{Feature Distillation at Classification Head.} 
Feature distillation can effectively mitigate forgetting caused by feature drift. However, since the feature layers also contain features of new-class objects, distilling all feature layers suppresses learning new-class features, thereby reducing the detector's plasticity. To address this, we distill only the positions of the generated objects. Specifically, the feature distillation can be expressed as:
\begin{equation}
{\mathcal{L}_{dist - feat}} = \sum\limits_{b\in B_{r}} {{\mathcal{L}_2}(RA({\mathcal{F}_{ori}},b),RA({\mathcal{F}_{inc}},b))} 
\label{eq6}
\end{equation}
where $\mathcal{F}_{ori}$ and $\mathcal{F}_{inc}$ are the last feature layer from the classification head of the original and incremental detectors. $RA$ is the ROIAlign proposed by \cite{he2017mask}, which extracts the generated objects' features based on the image's replayed bounding boxes $B_{r}$.

\textbf{Logit Distillation at Classification and Regression Head.} In GFLV1, the logit output of the classification head is modeled as probabilities for object categories, and the logit output of the regression head is modeled as the deterministic for object boundaries, resulting in the position that having a high response indicates a higher likelihood of the presence of a generated or real object. Therefore, we transfer knowledge about old-class objects to the incremental detector by distilling the top-K high-response logit outputs in the original detector. Specifically, the logit distillation can be expressed as:
\begin{equation}
{\mathcal{L}_{dist - logit}} = \sum\limits_{j = 1}^k {\mathcal{L}_2(\mathcal{C}_{ori}^j,\mathcal{C}_{inc}^j) + {\mathcal{L}_{KL}}(\mathcal{B}_{ori}^j,\mathcal{B}_{inc}^j)}
\label{eq7}
\end{equation}
where $\mathcal{C}_{ori}^j$ and $\mathcal{C}_{inc}^j$ are the corresponding top-K selected responses from the classification heads of the original and incremental detectors; $\mathcal{B}_{ori}^j$ and $\mathcal{B}_{inc}^j$ are the corresponding top-K selected responses from the regression heads of the original and incremental detectors. Notably, to reduce the compression of category information by the sigmoid activation in the classification head, we use the $\mathcal{L}_2$ loss to distill the responses without the activation function. Since the regression responses in GFLV1 are represented as probability distributions, we use KL-Divergence as the distillation loss ${\mathcal{L}_{KL}}$ for the regression head.

\section{EXPERIMENTS}
\label{sec: Experiments}

\subsection{Experimental Settings}

We evaluate the proposed method on the publicly available dataset MS COCO 2017 \cite{lin2014microsoft}. We set up two versions, the non co-occurrence and the co-occurrence scenarios. In the non co-occurrence scenarios, images with both old and new class objects are only put into the original dataset, ensuring that old class objects are absent in the incremental dataset. In the co-occurrence scenarios, such images are put into both datasets, allowing old class objects to be present in the incremental dataset. We use GFLV1 \cite{li2020generalized} as the fundamental detector, utilizing a pre-trained ResNet-50 \cite{he2016deep} as its backbone, with batch norm layers \cite{ioffe2015batch} frozen during training. The optimizer is set up as the original paper \cite{li2020generalized}. All experiments are performed on an NVIDIA RTX 4090 with a batch size of 8.

\begin{table}[th]
\centering
\caption{Average precision (\%) for one-step comparisons with other methods on MS COCO 2017 dataset.}
\label{tab:One-step Comparisons}
\begin{adjustbox}{max width=\textwidth}
\setlength{\tabcolsep}{4pt} % Adjust the column separation here
\begin{tabular}{c|c|c|c|c|c}
\toprule
Setting & Method & 40+40 & 50+30 & 60+20 & 70+10 \\ 
\midrule
\multirow{6}{*}{Co-occurrence} & Upper Bound & 39.0 & 39.0 & 39.0 & 39.0 \\ 
\cmidrule(r){2-6} 
 & RILOD \cite{li2019rilod} & 29.9 & 28.5 & 25.4 & 24.5 \\
 & SID \cite{peng2021sid} & 34.0 & 33.8 & 32.7 & 32.8 \\ 
 & ERD \cite{feng2022overcoming} & 36.9 & 36.6 & 35.8 & 34.9 \\ 
 & PesudoRM \cite{yang2023pseudo} & 25.3 & - & - & - \\ 
 & IOR (our) &35.5 & 36.6 & 36.3& 35.9 \\
\midrule
\multirow{6}{*}{Non Co-occurrence} & Upper Bound & 34.4 & 34.1 & 34.2 & 36.6 \\ 
\cmidrule(r){2-6} 
 & RILOD \cite{li2019rilod} & 18.5 & 20.9 & 18.1 & 15.6 \\
 & SID \cite{peng2021sid} & 22.8 & 25.1 & 23.7 & 22.3\\ 
 & ERD \cite{feng2022overcoming} & 27.2 & 27.3 & 26.9 & 25.5 \\
 & PesudoRM \cite{yang2023pseudo} & 24.7 & - & - & - \\
 & IOR (our) & \textbf{30.3} & \textbf{30.1} & \textbf{29.1} & \textbf{29.9} \\
\bottomrule
\end{tabular}
\end{adjustbox}
\vspace{-7pt} 
\end{table}

\begin{table}[th]
\centering
\caption{Average precision (\%) for multi-step comparisons with other methods on MS COCO 2017 dataset.}
% (1-40) is the one-step normal training for categories 1-40 and +(a-b) is the incremental training for categories a-b.
\label{tab:Multy-step Comparisons}
\begin{adjustbox}{max width=\textwidth}
\setlength{\tabcolsep}{1pt} % Adjust the column separation here
\begin{tabular}{c|c|c|c|c|c|c}
\toprule
% Setting & Method & A(1-40) & +B(40-50) & +B(50-60) & +B(60-70)& +B(70-80) \\ 
Setting & Method & (1-40) & +(40-50) & +(50-60) & +(60-70)& +(70-80) \\ 
\midrule
\multirow{4}{*}{Co-occurrence} & RILOD \cite{li2019rilod} & \multirow{4}{*}{ 45.7 } & 25.4 & 11.2 & 10.5 & 8.4 \\
 & SID \cite{peng2021sid} & & 34.6 & 24.1 & 14.6 & 12.6 \\ 
 & ERD \cite{feng2022overcoming} & & 36.4 & 30.8 & 26.2 & 20.7 \\ 
 & IOR (our) & & 36.9 & 31.2 & 26.5 & 21.3 \\ \midrule
\multirow{4}{*}{Non Co-occurrence} & RILOD \cite{li2019rilod} & \multirow{4}{*}{ 45.7 } & 19.1 & 9.9 & 8.3 & 6.9  \\ 
 & SID \cite{peng2021sid} & & 27.9 & 18.3 & 12.5 & 9.6 \\ 
 & ERD \cite{feng2022overcoming} & &31.1 & 22.9 & 18.0 & 14.2  \\
 & IOR (our) & & \textbf{33.9} & \textbf{25.1} & \textbf{20.7} & \textbf{15.7}  \\
\bottomrule
\end{tabular}
\end{adjustbox}
\vspace{-15pt}
\end{table}

\subsection{Comparison Experiment}

As shown in Table \ref{tab:One-step Comparisons} and \ref{tab:Multy-step Comparisons}, we compare IOR with distillation-based methods, including RILOD \cite{li2019rilod}, SID \cite{peng2021sid}, ERD \cite{feng2022overcoming}, and the generation-based method PesudoRM \cite{yang2023pseudo} in one-step and multi-step setting. IOR achieves state-of-the-art performance in all non co-occurrence scenarios, significantly improving performance. Compared to distillation-based methods, IOR generates old-class objects for distillation, thereby alleviating catastrophic forgetting. Compared to PesudoRM \cite{yang2023pseudo}, which generates old-class samples through additional generators, IOR inverses the detector, which generates distinctive features and eliminates the requirement for additional generator training. In this way, IOR efficiently compensates for the absence of old-class objects in non co-occurrence scenarios. Moreover, IOR also shows good performance in the co-occurrence scenarios.

\begin{table}[th]
\centering
\caption{Ablation study under "40+40" setting in non co-occurrence scenarios.}
\label{tab:Ablation study1}
% \begin{adjustbox}{max width=\textwidth}
\setlength{\tabcolsep}{4pt} % Adjust the column separation here
\begin{tabular}{ccc|ccc}
\toprule
\multirow{2}{*}{Distillation} & Augmented & High-value & Old-class & New-class & Total \\ 
 & Replay & Distillation & AP & AP & AP \\ 
\midrule
$\checkmark$ & & & 33.7 & 20.0 & 26.9 \\ 
$\checkmark$ & $\checkmark$ & & 33.3 & 20.1 & 26.7 \\ 
$\checkmark$ & $\checkmark$ & $\checkmark$ & 40.2 & 20.3 & 30.3 \\ 
\bottomrule
\end{tabular}
\vspace{-10pt} 
% \end{adjustbox}
\end{table}

\subsection{Analysis and Ablation Study}
\label{sec:Analysis and Ablation Study}
Table \ref{tab:Ablation study1} evaluates each component on the "40+40" setting in non co-occurrence scenarios. The method with only distillation exhibits poor accuracy due to the lack of old-class objects. After replaying the generated objects, there is no significant change in accuracy because the background overwhelms the generated objects. When high-value distillation is added, the old-class AP significantly increases by 6.5$\%$, which indicates that the combination of components can effectively compensate for the absence of old-class objects.

\begin{table}[th]
  \centering
        \caption{Varing amounts of generated objects.}
        \begin{tabular}{c|ccc}
        \toprule
        \multirow{2}{*}{Amount} & Old-class & New-class & Total \\ 
         & AP & AP & AP \\ 
        \midrule
        1 & 39.2      &  20.6     & 29.9      \\
        3 &  39.3     &  20.3     & 29.8     \\
        10 & 40.2     &  20.3     &  30.3     \\
        30 & 39.9     &  20.4     &  30.2     \\
        100 & 39.6    &  20.5     &  30.0     \\
        \bottomrule
        \end{tabular}
    \label{tab: Number of Generated Samples}
    \vspace{-15pt}
\end{table}

\begin{table}[th]
\centering

% \captionsetup{aboveskip=5pt, belowskip=10pt}
\caption{Effect of different inversing loss.}
\label{tab:Effect of different inversing loss}
\begin{tabular}{ccc|ccc}
\toprule
\multirow{2}{*}{${\mathcal{L}_{detect}}$} & \multirow{2}{*}{${\mathcal{R}_{BN}}$} & \multirow{2}{*}{${\mathcal{R}_{prior}}$} & Old-class & New-class & Total \\ 
  & & & AP & AP & AP \\ 
\midrule
$\surd$ & & & 39.5 & 19.9 & 29.7 \\ 
$\surd$ & $\surd$ & & 39.7 & 20.1& 29.8\\ 
$\surd$ & & $\surd$ & 39.8& 19.9& 29.9\\ 
$\surd$ & $\surd$ & $\surd$ & 40.2  &  20.3 & 30.3 \\ 
\bottomrule
\end{tabular}
\end{table}
\textbf{Requirement of Generated Object.} Table \ref{tab: Number of Generated Samples} lists the results under different amounts of generated objects. We observe that the maximum total AP gap is merely 0.5$\%$, which indicates that generating few objects can efficiently complement the absence of old-class objects. Thus, the proposed method can improve the IOD performance efficiently.

\textbf{Effect of Inversing Loss.} Table \ref{tab:Effect of different inversing loss} lists the effect of different inversing losses. After removing ${\mathcal{R}_{BN}}$ and ${\mathcal{R}_{prior}}$, the total AP only decreases by 0.6\%, which indicates that the category information provided by ${\mathcal{L}_{detect}}$ contributes more to accuracy.

% \textbf{Actual Generation Time.} In the "40+40" setting, generating one object per class takes only 10 minutes on an NVIDIA GeForce RTX 4090. Compared to 12 hours of incremental training, this additional 10 minutes is negligible. The generated objects result in a 5.5\% increase in old-class accuracy and a 3.0\% increase in overall accuracy, demonstrating the efficiency of IOR.

\textbf{Efficiency Comparison.} Since the previous generation-based methods do not release code, a detailed comparison of total consumption is infeasible. SDDGR \cite{kim2024sddgr} introduces pre-trained CLIP \cite{radford2021learning} and SD \cite{rombach2022high}, requiring about 5GB of extra storage. PesudoRM \cite{yang2023pseudo} trains a generative model, converging after approximately 100,000 iterations. Furthermore, both methods generate extensive samples, resulting in high generation costs. In contrast, IOR has no storage and iteration consumptions for generative models. In sample generations, it generates one object per category in just 10 minutes under the "40+40" setting using an NVIDIA RTX 4090, which improves total AP by 3.0\% compared to the baseline. This 10-minute generation time is negligible compared to 12 hours of incremental training, indicating IOR's efficiency.

\section{Conclusion}
In this paper, we propose Inversed Objects Replay (IOR) to efficiently mitigate the accuracy degradation of distillation-based IOD in non co-occurrence scenarios. To efficiently complement the absence of old-class objects, we generate old-class objects by inversing the original detector without the requirements of training and storing extra generative models. Furthermore, we propose the augmented replay to reuse the generated objects, thus reducing the generation requirements. Finally, we propose the high-value distillation method to mitigate the background interference during the distillation.

% \section*{References}
\clearpage
\printbibliography

\end{document}